\documentclass[conference]{IEEEtran}
\IEEEoverridecommandlockouts 

\usepackage{cite}

\usepackage{lipsum}
\usepackage{xcolor}
\usepackage{graphicx}
\usepackage[all]{nowidow}
\usepackage{tikz,hyperref}
\usepackage{microtype}
\usepackage{makecell}
\usepackage{comment}
\usepackage{multirow}
\usepackage[caption=false,font=normalsize,labelfont=normalfont,textfont=normalfont]{subfig}
\ifCLASSINFOpdf
\else
\fi
\begin{document}
%
\title{Cross or Wait? Predicting Pedestrian Interaction Outcomes at Unsignalized Crossings
\thanks{This research is funded by the European research project ``SHAPE-IT – Supporting the Interaction of Humans and Automated Vehicles: Preparing for the Environment of Tomorrow’’. This project has received funding from the European Union’s Horizon 2020 research and innovation programme under the Marie Skłodowska-Curie grant agreement 860410. \\
This paper has been accepted for publication in 2023 IEEE Intelligent Vehicles Symposium (IV): \url{https://doi.org/10.1109/IV55152.2023.10186616} \copyright 2023 IEEE

}
}

\author{\IEEEauthorblockN{Chi Zhang\IEEEauthorrefmark{1},
Amir Hossein Kalantari\IEEEauthorrefmark{2},
Yue Yang\IEEEauthorrefmark{2}, 
Zhongjun Ni\IEEEauthorrefmark{3}, \\
Gustav Markkula\IEEEauthorrefmark{2},
Natasha Merat\IEEEauthorrefmark{2}, and
Christian Berger\IEEEauthorrefmark{1}} 
\IEEEauthorblockA{\IEEEauthorrefmark{1} Department of Computer Science and Engineering, University of Gothenburg, Sweden.\\ Email: \{chi.zhang, christian.berger\}@gu.se}
\IEEEauthorblockA{\IEEEauthorrefmark{2} Institute for Transport Studies, University of Leeds, Leeds LS2 9JT, UK. 
}
\IEEEauthorblockA{\IEEEauthorrefmark{3} Department of Science and Technology, Link{\"o}ping University, Campus Norrk{\"o}ping, Sweden. 
}
}

\maketitle


\begin{abstract}
Predicting pedestrian behavior when interacting with vehicles is one of the most critical challenges in the field of automated driving. Pedestrian crossing behavior is influenced by various interaction factors, including time to arrival, pedestrian waiting time, the presence of zebra crossing, and the properties and personality traits of both pedestrians and drivers. However, these factors have not been fully explored for use in predicting interaction outcomes.
In this paper, we use machine learning to predict pedestrian crossing behavior including pedestrian crossing decision, crossing initiation time (CIT), and crossing duration (CD) when interacting with vehicles at unsignalized crossings. Distributed simulator data are utilized for predicting and analyzing the interaction factors. Compared with the logistic regression baseline model, our proposed neural network model improves the prediction accuracy and F1 score by 4.46\% and 3.23\%, respectively. Our model also reduces the root mean squared error (RMSE) for CIT and CD by 21.56\% and 30.14\% compared with the linear regression model. Additionally, we have analyzed the importance of interaction factors, and present the results of models using fewer factors. This provides information for model selection in different scenarios with limited input features.
\end{abstract}


%
\IEEEpeerreviewmaketitle

\section{Introduction}
The demand for protecting pedestrians impels the vehicular automation industry to develop automated driving (AD) technologies. One of the most critical and challenging tasks in this domain is understanding and predicting pedestrian behavior during crossing, especially when they are interacting with vehicles at unsignalized crossings. Unsignalized crossings are crossings without signal displays or traffic lights, and can be marked (zebra crossings) or unmarked (non-zebra crossings). The interaction outcome prediction can provide AD systems and drivers with additional information to make safer decisions, thereby preventing vehicle-pedestrian conflicts.

Predicting pedestrian crossing behavior is challenging because pedestrian behavior is influenced by many factors~\cite{Rasouli2019Autonomous}.
Many researchers have investigated the objective interaction factors that impact pedestrian-vehicle interaction such as time to arrival (TTA), waiting time, crossing location, age, and gender~\cite{gorrini2018observation,velasco2021will,domeyer2020interdependence,Kalantari2022Who,wang2021investigating,domeyer2022driver,cloutier2017outta,habibovic2018communicating,Rasouli2019Autonomous}, and the subjective interaction factors related to pedestrians' personality traits such as sensation seeking (SS) and social value orientation (SVO) for prediction~\cite{rosenbloom2006sensation,crosato2021human,Kalantari2022Who}. These studies have investigated factors that impact interaction and crossing behavior. However, they only focused on analyzing the effects of the interaction from a statistical point of view, but have not evaluated the predictability of the pedestrian crossing models with these factors.

Some studies predicted crossing intention or action using machine learning-based methods and neural networks~\cite{Fang2018,Gujjar2019Classifying,Chaabane2020,yang2021crossing,zhang2021pedestrian}, but they mainly considered pedestrians' states and environment cues, while neglecting the interaction between pedestrians and vehicles. This may be because these studies used naturalistic data making it hard to study interaction factors without being cautious about potential latent variables that might affect pedestrian behavior. Furthermore, it is usually difficult to distinguish the \textit{original intention} from the \textit{actual action} in naturalistic data, thereby making it intractable to investigate the outcome of the interaction. Moreover, it is impractical to get pedestrians' personality traits such as SS and SVO from naturalistic data.
Hence, a simulator study with controlled conditions is a way to understand pedestrian-vehicle interactions and develop predictive models.

To the best of the authors' knowledge, there are no studies on machine learning-based predictive models for pedestrian crossing behavior that focus on pedestrian-vehicle interaction at unsignalized crossings and consider personality traits. To fill this research gap, we use the dataset collected from a distributed simulator to investigate pedestrian-vehicle interaction while crossing and predict pedestrian crossing behavior.
In this paper, we aim to develop machine learning-based models to predict whether a pedestrian will cross first or wait when interacting with a vehicle at unsignalized crossings. We consider both objective factors related to crossing conditions and subjective factors related to personalities, and develop predictive models using machine learning methods. The key factors that influence pedestrian crossing decisions are analyzed and identified.
The main contributions of this paper are:
\begin{enumerate}
    \item We use the distributed simulator data collected by Kalantari et al.~\cite{Kalantari2022Who} for prediction and analysis. We build a baseline predictive model using logistic regression and linear regression with the factors used by Kalantari et al. Compared with previous statistical models, our predictive model provides the baseline information on the predictability of pedestrian interaction outcomes.

    \item We develop machine learning interaction models for prediction using the support-vector machine~(SVM), random forest~(RF), and neural network~(NN). We investigate factors including TTA, pedestrian waiting time, being at zebra crossings or not, and both the driver's and the pedestrian's age, gender, SVO slider measure~\cite{murphy2011measuring}, and Arnett inventory of sensation seeking (AISS)~\cite{arnett1994sensation}. Compared to the baseline model, our proposed crossing decision model improves the accuracy and F1 score by 4.46\% and 3.23\%. Our model also reduces the mean absolute error (MAE) and root mean squared error (RMSE) by 30.84\% and 21.56\% for crossing initiation time (CIT) prediction, and by 35.00\% and 30.14\% for crossing duration (CD) prediction.
    
    \item We analyze interaction factors that influence pedestrian crossing behavior. We do an ablation study to investigate how a model performs when lacking the information of drivers' and pedestrians' age, gender, SVO, and AISS information. This provides guidance on model selection in scenarios when there are only partial input features.
\end{enumerate}

\section{Related Work}
\subsection{Studies on Pedestrian-Vehicle Interaction}
Studies have investigated factors that impact pedestrian-vehicle interactions~\cite{Rasouli2019Autonomous,kotseruba2020they}. These interaction factors include time-related factors such as time gap or time to arrival (TTA)~\cite{gorrini2018observation, velasco2021will,Kalantari2022Who} and waiting time~\cite{wang2021investigating,domeyer2022driver,Kalantari2022Who}, the vehicle's dynamics including velocity and deceleration~\cite{Volz2016}, crossing places such as whether at zebra crossings or not~\cite{cloutier2017outta,habibovic2018communicating,Kalantari2022Who}, and demographics such as age and gender~\cite{gorrini2018observation, Rasouli2019Autonomous, velasco2021will, cloutier2017outta,Kalantari2022Who}.

In addition to these objective factors, psychological subjective factors that reflect drivers' and pedestrians' individual personality traits also influence their decision-making during the interaction. Arnett~\cite{arnett1994sensation} defined SS as the need for varied and novel experiences, and the inclination to take risks for such experiences. The author presented a new scale namely the Arnett inventory of sensation seeking (AISS) that is related to risk behavior. SS has been found to be related to pedestrian crossing behavior~\cite{rosenbloom2006sensation,wang2022effect,Kalantari2022Who}. Individuals with higher SS are more likely to take risks in traffic. SVO is used to describe how much an individual values the welfare of others compared to their own, and the individual's preference for distributing resources in social exchanges~\cite{mcclintock1989social,murphy2011measuring,murphy2014social}, and can be measured by SVO slider measure~\cite{murphy2011measuring}. SVO is found to be related to a person's decision making~\cite{crosato2021human,crosato2022interaction,Kalantari2022Who}. Individuals with higher SVO values are more likely to be altruistic and yield or wait while interacting in traffic, while those with lower SVO values tend to behave more aggressively.

However, most of these studies are from a statistical view, while less attention has been paid to predictive models to evaluate the accuracy that can be achieved using these factors. To contribute to the AD industry, only understanding the influencing factors of interaction outcomes is not sufficient. In this paper, we build predictive models to provide additional information to AD systems.  

\subsection{Pedestrian Crossing Behavior Prediction}
Many existing learning-based models on pedestrian crossing behavior prediction are based on naturalistic data~\cite{Fang2018,Gujjar2019Classifying,Chaabane2020,yang2021crossing,zhang2021pedestrian}. These studies mainly focused on the features of the pedestrians and the environment, but have not considered the interaction between pedestrians and vehicles explicitly. One possible reason is that these studies used naturalistic data collected under uncontrolled conditions. Since there are too many latent factors that can simultaneously affect pedestrian behavior and pedestrian-vehicle interaction in naturalistic data, it is difficult to investigate the interactions. Furthermore, without a controlled environment, it is hard to distinguish the \textit{intention} and the \textit{actual action} in naturalistic data, as labeling the intention of a pedestrian is a challenging task. However, without knowing the intention, it is hard to analyze the outcome of the interaction. Rasouli et al.~\cite{Rasouli2019PIE} addressed and labelled intention by asking multiple annotation participants to observe the video of pedestrians and label the crossing intention, and then took the average, which is very time-consuming and costly. Besides, the naturalistic data contain noise making it harder to identify the factors that influence interactions.

Moreover, although some of the studies using naturalistic data considered pedestrian-vehicle interaction~\cite{Volz2015,Volz2016,Zhang2020Research}, it is hard to get the human factor parameters and personality traits such as SVO slider measure and AISS without the help of questionnaires. V{\"o}lz et al.~\cite{Volz2015,Volz2016} predicted pedestrian crossing behavior at zebra crossing considering the relative position and relative velocity between a pedestrian and a vehicle. Zhang et al.~\cite{Zhang2020Research} proposed a predictive model at the zebra crossing, considering factors including TTA, the speed and position of both the vehicle and the pedestrian, and the age and gender of the pedestrian.
Jayaraman et al.~\cite{jayaraman2020analysis} used simulator data for analyzing and predicting the crossing behavior considering waiting time, but they only focused on interaction with AV approaching at different speeds.
These studies have not included personality traits such as SVO and AISS due to the scarcity in such information, and they also have not investigated the influence of the presence of zebra crossing.
Therefore, to overcome the challenges of using naturalistic data for crossing behavior prediction, this paper is based on a distributed simulator study (DSS). We focus on pedestrian-vehicle interaction and consider personality traits and the presence of zebra crossing. 

\begin{figure*}[!t]
\centering
\subfloat[]{\includegraphics[width=0.45\textwidth]{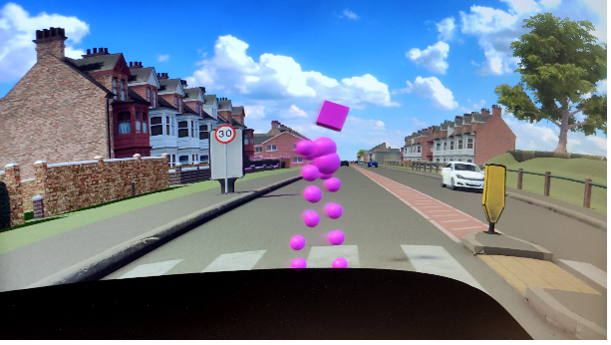}%
\label{fig:data_driver_view}}
\hfil
\subfloat[]{\includegraphics[width=0.45\textwidth]{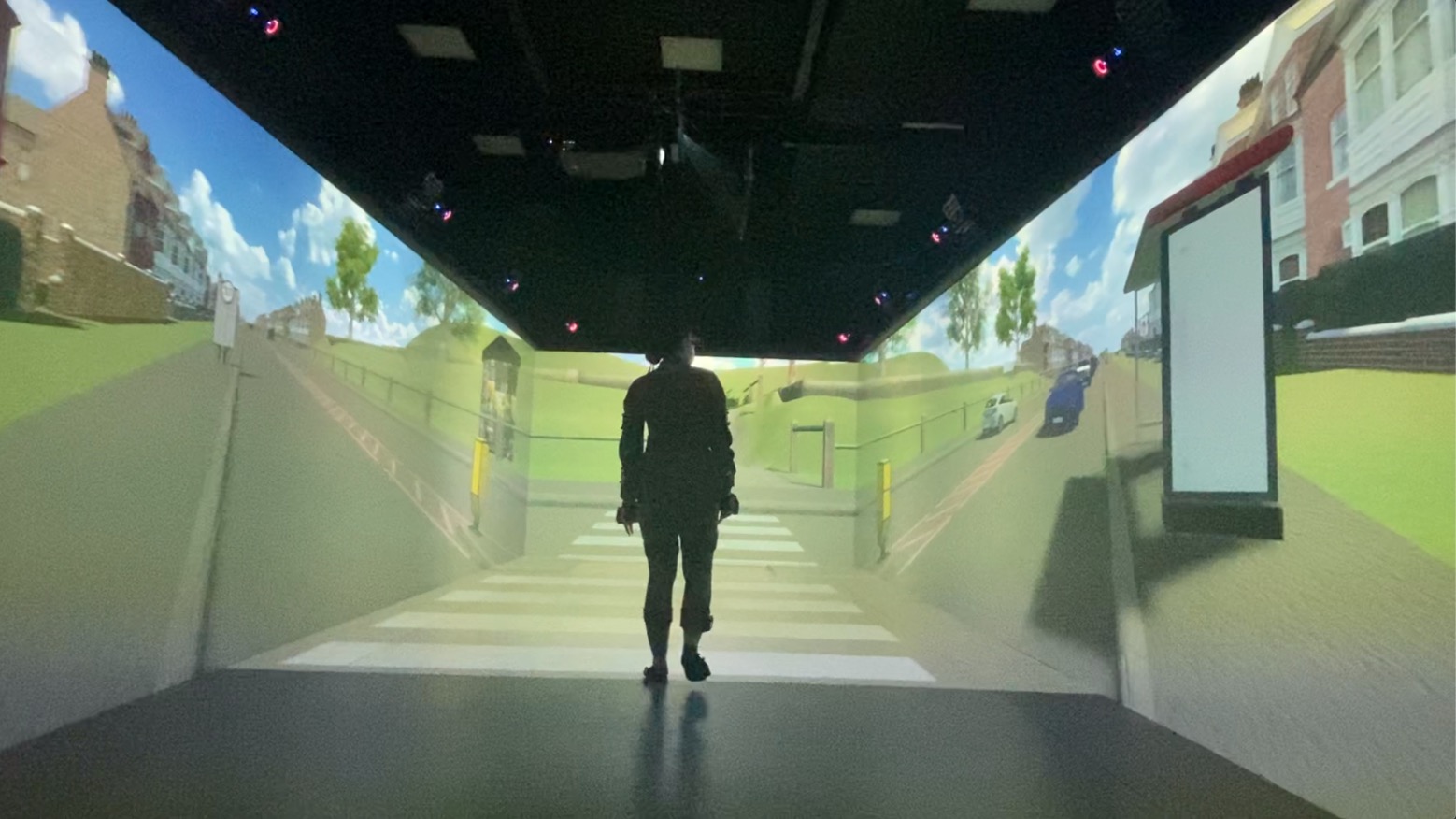}%
\label{fig:data_pedestrian_view}}
\caption{Illustration for the distributed simulator study (DSS). (a) A pedestrian from the driver’s view: the pink bubbles are the body markers representing the pedestrian. (b) An interaction example: a pedestrian is crossing the zebra and interacting with the vehicle to their right.}
\label{fig:data_participants_view}
\end{figure*}

\begin{figure*}
    \centering
    \includegraphics[width=0.9\textwidth]{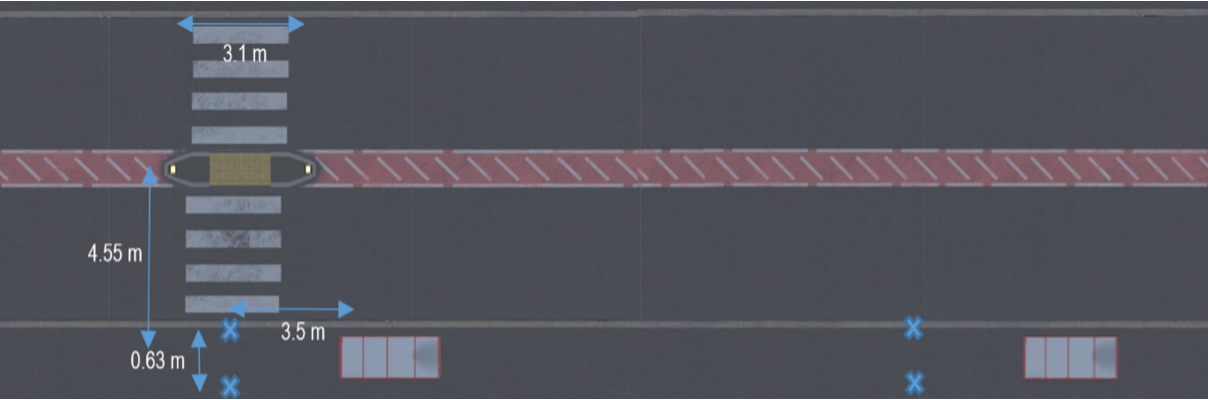}
    \caption{Top view of the zebra (left) and non-zebra crossing (right) with the designated standpoints (blue $\times$ markers). The first marker shows the pedestrian’s standing point, and the second shows the curb of the virtual road. The grey rectangles are visual obstructions (bus stops).}
    \label{fig:data_top_view}
\end{figure*}

\section{Methodology}

\subsection{Distributed Simulator Data Collection}
In this section, we briefly introduce the dataset. We use the same data collected by Kalantari et al.~\cite{Kalantari2022Who} in their distributed simulator study (DSS).

The DSS was conducted by connecting two high-fidelity simulators: the University of Leeds Driving Simulator (UoLDS) was connected to the Highly Immersive Kinematic Experimental Research (HIKER) pedestrian lab. UoLDS is a motion-based driving simulator having eight degree-of-freedom which is housed in a 4 m-diameter spherical projection dome. The field-of-view of the projection system is 300°. HIKER is a CAVE-based simulator with dimensions equal to $9\times4$~m and eight Barco F90 4k projectors used to project virtual scenes at 120~Hz to its floor and walls. Fourteen body markers were attached to the pedestrian’s body representing them as pink spheres to the driver as shown in Fig.~\ref{fig:data_driver_view}. The vehicle was also observable as an entity to the pedestrian as shown in Fig.~\ref{fig:data_pedestrian_view}. 

Sixty-four participant pairs (32 drivers at age: mean $(M)=31.53$, range $(R)=21-50$, standard deviation $(SD)=1.72$; paired with 32 pedestrians at age: $M=25.09$, $R=19-34$, $SD = 0.87$) interacted with each other under different scenarios. The traffic scenarios relied on five different TTAs (3, 4, 5, 6 and 7~s) and four crossing locations (two zebra and two non-zebra crossings as shown in Fig.~\ref{fig:data_top_view}). This led to 20 conditions repeated in two separate experimental blocks, resulting in 40 randomized trials per participant pair. An 890~m two-way urban road with traffic on both lanes was created in Unity 3D to apply these settings.

Both participants were told that they were with the assumption that they were in a hurry (e.g., late for a work meeting) while they wanted to pass safely as well. They were also reminded that the priority at zebra crossings was for pedestrians. The pedestrian was asked to stand at a point on the HIKER’s floor where their vision was obstructed by a bus stop on the right side making them unable to see when the subject vehicle was approaching but they could see that vehicles were going both ways. Whenever they heard an auditory tone, they needed to step up to a second point where they could see the subject vehicle, and then they could decide whether to cross or not.
Drivers were told that they needed to watch the speed limit (30~mph) and they could see that sometimes pedestrians are coming from behind one of these bus stops. Likewise, they could either drive on or slow down and yield to the pedestrians.

\subsection{Input and Output Variables}
\label{sec:problem and variables}
We aim to predict the ``cross or wait'' decisions of all pedestrians, and for those crossing cases, we are also concerned about their crossing initiation time (CIT) and crossing duration (CD).
The candidate variables to be used as prediction input features are listed in Table~\ref{tab:independent_variables}. 
Kalantari et al.~\cite{Kalantari2022Who} used the participant pair as a feature. However, as we usually cannot get the same participant pair in a real driving scenario, i.e., the same driver meets the same pedestrian, this information is not practical to be used for prediction. So we do not use participant pair as our input feature.
The interaction outcomes are listed in Table~\ref{tab:dependent_variables}. 

\begin{table}[h]
    \caption{Candidate variables used for prediction (Model Inputs).}
    \label{tab:independent_variables}
    \centering
    \begin{tabular}{m{2cm}|m{6cm}}
    \hline
        Variable [Unit] & Description (type) \\
    \hline
        $T_{a}$ [s] & Time to arrival (TTA), which was set to 3, 4, 5, 6, and 7~s. TTA represents the time for the vehicle to approach the center of the crossing (continuous) \\ \hline
        $T_{w}$ [s] & Waiting time ($M=52.71$, $R=13.8-106.98$, $SD=19.04$), the time that the pedestrian needed to wait in each trial before prompting to cross by auditory tone. In the actual environment, it denotes the pedestrians' duration of stay towards the crossing area due to different reasons such as working with their smartphones (continuous) \\ \hline
        L & Crossing location type, including two categories: zebra and non-zebra (categorical) \\ \hline
        $A_d$, $A_p$ & Age for both pedestrians and drivers. (discrete) \\ \hline 
        $G_d$, $G_p$ & Gender for both pedestrians and drivers (categorical) \\ \hline
        $SVO_d$, $SVO_p$ [degree] & SVO slider measure for both pedestrians and drivers, calculated from the SVO slider measure questionnaire~\cite{murphy2011measuring}. $SVO_d$: $M=53.17$, $R=45.00-78.38$, $SD=8.35$, $SVO_p$: $M=53.67$, $R=43.92-75.26$, $SD=7.82$ (continuous) \\  \hline%
        $AISS_d$, $AISS_p$ & AISS for both pedestrians and drivers, calculated from the 20-item AISS questionnaire~\cite{arnett1994sensation}. $AISS_d$: $M=53.78$, $R=43.00-69.00$, $SD=6.70$, $AISS_p$: $M=50.47$, $R=27.00-61.00$, $SD=7.17$ (continuous) \\
    \hline
    \end{tabular}
\end{table}

\begin{table}[h]
    \caption{Variables of pedestrian interaction outcomes (Model Outputs).}
    \label{tab:dependent_variables}
    \centering
    \begin{tabular}{l|m{6cm}}
    \hline
         Variable [Unit] & Description (type) \\
    \hline
         P & Cross or wait, pedestrians' crossing decision as outcome of the interaction, 1 for crossing, 0 for waiting (binary categorical) \\ \hline
         CIT [s] & Crossing initiation time, from the time the pedestrians were prompted to cross by the auditory tone to the time they started crossing the road (continuous)  \\ \hline
         CD [s] & Crossing duration, the time pedestrians started crossing to the time they reached the central hatch (continuous)  \\
    \hline
    \end{tabular}
\end{table}

Three feature sets are used in prediction, as listed below.

\begin{itemize}
    \item \textbf{Baseline:} features used in Kalantari et al.'s work~\cite{Kalantari2022Who}, including $T_{a}$, $T_{w}$, $L$, $A_p$, $G_p$, $\Delta SVO$, $\Delta AISS$, $Participant Pair$, where $\Delta SVO = SVO_p - SVO_d$, $\Delta AISS = AISS_p - AISS_d$.
    \item \textbf{Ours:} all input features listed in Table~\ref{tab:independent_variables}, including $T_{a}$, $T_{w}$, $L$, $A_d$, $A_p$, $G_d$, $G_p$, $SVO_d$, $SVO_p$, $AISS_d$, $AISS_p$.
    \item \textbf{Ours with $\Delta SVO$, $\Delta AISS$:} in addition to the listed input features, the delta information of SVO and AISS is added. The features include $T_{a}$, $T_{w}$, $L$, $A_d$, $A_p$, $G_d$, $G_p$, $SVO_d$, $SVO_p$, $AISS_d$, $AISS_p$, $\Delta SVO$, $\Delta AISS$.
\end{itemize}


\subsection{Predictive Models} 
The following machine learning models are used for predicting pedestrian-vehicle interaction.

\paragraph{Linear Models} Logistic regression and linear regression were used by Kalantari et al.~\cite{Kalantari2022Who} for analysis. We use them as a baseline for prediction.

\paragraph{Support-Vector Machine (SVM)} SVM aims to find a hyperplane in the feature space and can be used for classification. Here we use a linear kernel for classification, so the SVM is also considered as a linear-based model.

\paragraph{Random Forest (RF)} RF is an ensemble learning method for classification and regression. RF constructs a large number of decision trees. It outputs the most selected label for the classification task and outputs the average prediction for the regression task. The number of estimators is set to 100. The maximum depth is set to five to avoid over-fitting.

\paragraph{Neural Networks (NNs)} NNs are based on a collection of artificial nodes and can be used for both classification and regression. NNs usually contain several node layers, including an input layer, an output layer, and one or several hidden layers. As the number of our model's input features is small, here we use the multilayer perceptron (MLP), which is a fully connected feedforward NN.
Since the input variables are in different types and with different scales, normalization is required to scale the input variables in the same range. To avoid over-fitting of MLP models, we used two hidden layers with node sizes of 16 and four, respectively.

\subsection{Evaluation Metrics}
The prediction of the crossing decision is a classification problem, so we use prediction accuracy (ACC) and F1 score for evaluation. The evaluation functions are shown below, where P and N denote the numbers of positives and negatives, respectively. TP, TN, FP, FN are the numbers of true positives, true negatives, false positives, and false negatives, respectively.

\begin{equation}
ACC=\frac{TP+TN}{P+N}
\label{eq_acc}
\end{equation}

\begin{equation}
F1 =\frac{2TP}{2TP+FP+FN}
\label{eq_f1}
\end{equation}

The predictions of CIT and CD are regression problems, so we use mean absolute error (MAE) and root mean squared error (RMSE) for evaluation. The evaluation functions are as follows, where $y_i$ denotes the groundtruth for the $i^{th}$ trail, and $\hat y_i$ denotes the corresponding prediction, $n$ denotes the number of trails.
\begin{equation}
MAE =\frac{\Sigma|\hat y_i - y_i|}{n}
\label{eq_mae}
\end{equation}

\begin{equation}
RMSE =\sqrt{\frac{\Sigma(\hat y_i - y_i)^2}{n}}
\label{eq_rmse}
\end{equation}

\subsection{Model Implementation Details}
There were a total of 1279 trials collected, where 836 trials were crossing cases. To better evaluate models, we use five-fold cross-validation. We randomly divide the dataset into five sets, and each time we use one set for testing and the rest for training. There is no overlap between the training and test sets. We report the average performance over test sets.

The information of categorical features (e.g., crossing location type and gender) is represented by categories rather than in numeric formats. 
We apply the commonly used one-hot encoding to encode non-ordinal categorical variables.

\section{Results and Discussions}

\subsection{Crossing Decision}
\label{sec:results_crossing_decision}
The prediction accuracy and F1 score of the pedestrian crossing decision are shown in Table~\ref{tab:pred_acc}. 
Using our proposed MLP model, the accuracy and F1 score are improved by 4.46\% and 3.23\% compared with logistic regression based on the effective input features stated by Kalantari et al.~\cite{Kalantari2022Who}.

\begin{table}[h]
    \caption{Prediction Accuracy and F1 score of Crossing Decision}
    \label{tab:pred_acc}
    \centering
    \begin{tabular}{m{1.65cm}|m{0.7cm}m{0.7cm}|m{0.7cm}m{0.7cm}|m{0.7cm}m{0.7cm}}
    \hline
        Model & \multicolumn{2}{c|}{Zebra} & \multicolumn{2}{c|}{Non-zebra} & \multicolumn{2}{c}{Total} \\ 
        (features) & ACC & F1 & ACC & F1 & ACC & F1\\  \hline
        LR(baseline)~\cite{Kalantari2022Who} & 91.39\% & 95.32\% & 80.16\% & 75.81\% & 85.77\% & 89.24\% \\
        LR (ours) & 91.39\% & 95.30\% & 80.47\% & 76.01\% & 85.93\% & 89.32\% \\
        SVM (ours) & 91.24\% & 95.22\% & 81.09\% & 77.04\% & 86.16\% & 89.55\% \\
        RF (ours) & 91.24\% & 95.33\% & 88.44\% & 85.93\% & 89.84\% & 92.44\% \\
        MLP (ours) & 91.55\% & 95.28\% & \textbf{88.91\%} & \textbf{86.63\%} & \textbf{90.23\%} & \textbf{92.47\%} \\
    \hline
    \end{tabular}
\end{table}


We investigate the prediction results of zebra and non-zebra crossing (Table~\ref{tab:pred_acc}). The prediction results on zebra crossing cases are better than non-zebra crossing cases, showing that zebra crossing cases are more predictable.
All models perform similarly in zebra crossing cases. For non-zebra crossing cases, the non-linear models (RF and MLP) perform significantly better than the linear models (LR and SVM). Compared to the LR model with baseline features, our proposed MLP model improved the accuracy and F1 score by 8.75\% and 10.82\% respectively. This implies that the non-zebra crossing cases are more complex because of non-linearity and can be better handled by our proposed non-linear models.

\begin{figure}[!t]
\centering
\subfloat[]{\includegraphics[width=0.45\textwidth]{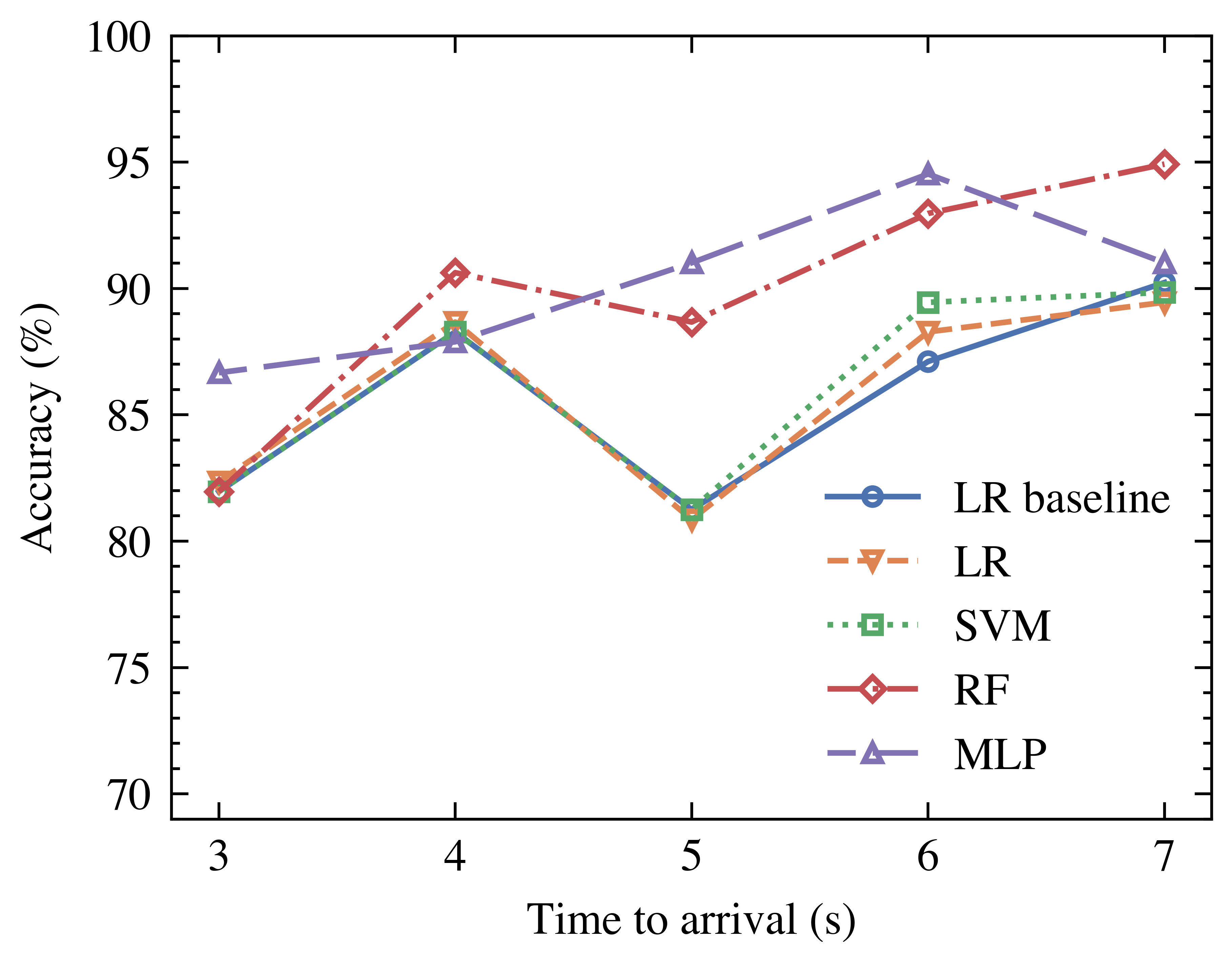}
\label{fig:acc_tta}}
\hfil
\subfloat[]{\includegraphics[width=0.45\textwidth]{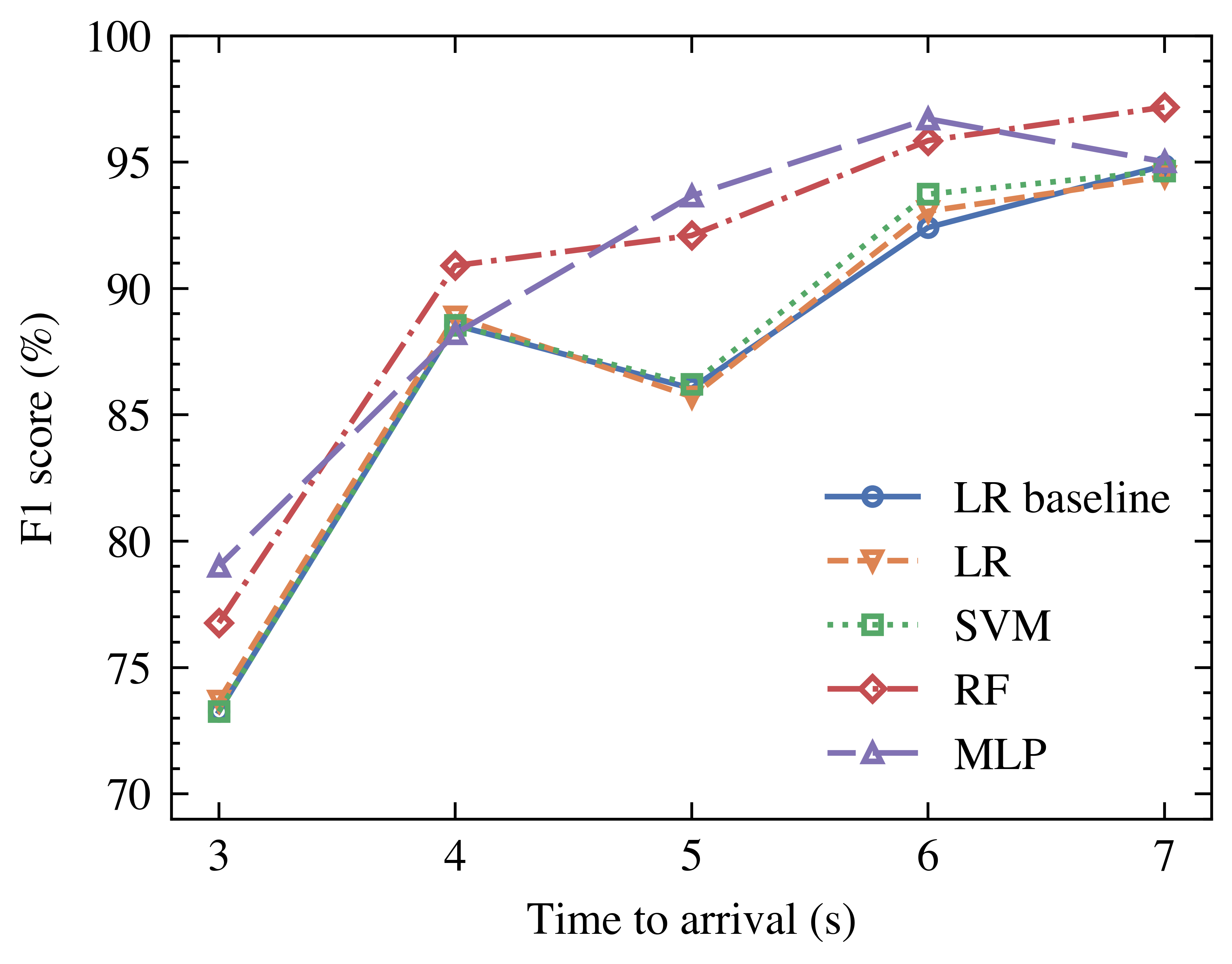}%
\label{fig:f1_tta}}
\caption{The (a) prediction accuracy and (b) F1 score versus time to arrival for logistic regression (LR), support-vector machine (SVM), random forest (RF), and multilayer perceptron (MLP) models.}
\label{fig:acc_and_f1_tta}
\end{figure}

The prediction accuracy and F1 score versus TTA for different models are shown in Fig.~\ref{fig:acc_and_f1_tta}. 
Smaller TTAs lead to lower prediction accuracy and F1 score, which indicates higher prediction difficulty. One possible reason is a smaller TTA means shorter reaction time and more interaction between pedestrians and drivers. This increases the uncertainty of crossing behavior and makes it harder to predict.
Non-linear models (RF and MLP) perform better than linear models (LR and SVM). This implies that the interaction process involves non-linearity that can be better handled by non-linear models.

The seven most important input features in each model are listed in Table~\ref{tab:feature_importance_classification}. As MLP is fully connected with all input features with hidden nodes, we do not list its feature importance here. In all three compared models, crossing location type $L$ (whether crossing at a zebra cross or not) and time to arrival $T_a$ are the most important features. The AISS of pedestrians and the pedestrian waiting time are also important for all three models. In linear models (LR and SVM), the models rely more on objective properties such as age and gender, while in non-linear models (RF), personality traits such as AISS and SVO are more important for prediction. As the non-linear models that rely on AISS and SVO for prediction gain better prediction results, this indicates that the personality traits contribute to the crossing decision prediction in a non-linear way. In Sec.~\ref{sec:ablation_study}, we present more results on prediction without these features to show the features' impacts.

\begin{table}[h]
    \caption{Most Important Features in the Crossing Decision Model}
    \label{tab:feature_importance_classification}
    \centering
    \begin{tabular}{l|l}
    \hline
        Model & Most Important Features (sorted by importance) \\  \hline
        LR & \makecell[l]{$L, T_{a}, G_p, AISS_p, A_d, A_p, T_w$} \\
        SVM & \makecell[l]{$L, T_{a}, G_p, AISS_p, A_d, T_w, SVO_p$}  \\
        RF & \makecell[l]{$T_{a}, L, T_{w}, AISS_d, SVO_d, AISS_p, A_p$} \\
    \hline
    \end{tabular}
\end{table}


\subsection{Crossing Initiation Time and Crossing Duration}
For crossing cases, we are also interested in the crossing initiation time (CIT) and crossing duration (CD) as interaction outcomes. The unit of MAE and RMSE is seconds. The smaller error represents better performance.

\paragraph{Crossing initiation time}
The prediction error of CIT is shown in Table~\ref{tab:pred_error_cit}. Features with $\Delta SVO$ and $\Delta AISS$ are used as inputs for RF and MLP models. Both RF and MLP show a great improvement compared with the baseline LR model, while RF achieves the best performance, and reduces the MAE and RMSE error by 30.84\% and 21.56\%, respectively.

\begin{table}[h]
    \caption{Prediction Error of Crossing Initiation Time}
    \label{tab:pred_error_cit}
    \centering
    \begin{tabular}{l|cc}
    \hline  
        Model (features) & MAE [s] & RMSE [s] \\  \hline
        LR (baseline)~\cite{Kalantari2022Who} & 0.618 & 0.897 \\
        RF (ours with $\Delta SVO$, $\Delta AISS$) &\textbf{0.428} & \textbf{0.704} \\
        MLP (ours with $\Delta SVO$, $\Delta AISS$) & 0.500 & 0.794 \\
    \hline
    \end{tabular}
\end{table}

\begin{figure}
    \centering
    \includegraphics[width=0.45\textwidth]{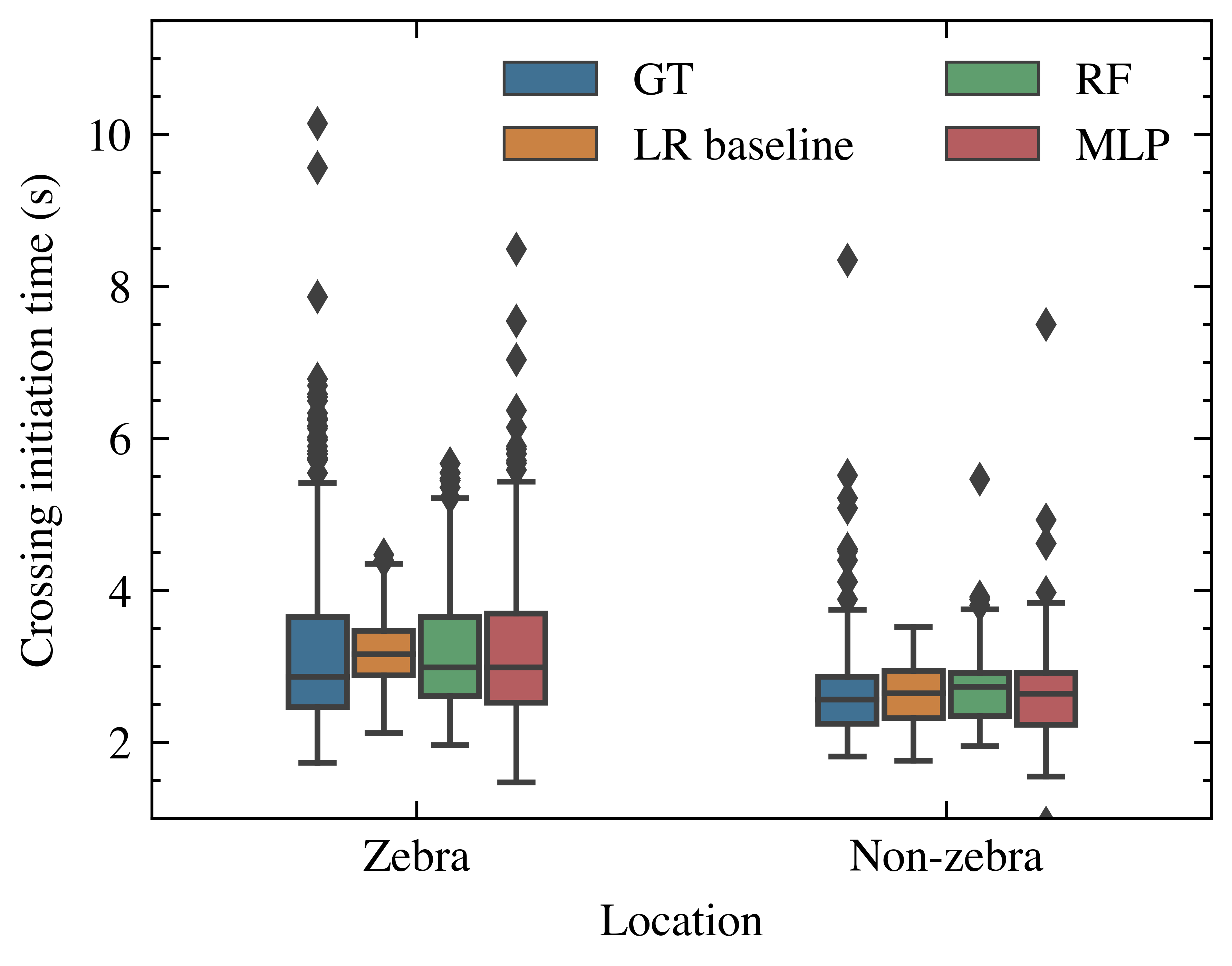}
    \caption{Box plots of crossing initiation time for groundtruth (GT), linear regression (LR), RF, and MLP models. Outliers are marked as diamond dots.}
    \label{fig:cit_boxplot}
\end{figure}

The box plots of CIT for the groundtruth and different models at different crossing locations are shown in Fig.~\ref{fig:cit_boxplot}. Compared with zebra crossing cases, non-zebra crossing cases have shorter CIT with a narrower distribution. This implies that when pedestrians cross at the non-zebra crossing, there are fewer chances for them to hesitate and they have to make a quicker decision. The prediction of the LR baseline model is more concentrated, while the MLP model is more distributed. 

\begin{figure}
    \centering
    \includegraphics[width=0.45\textwidth]{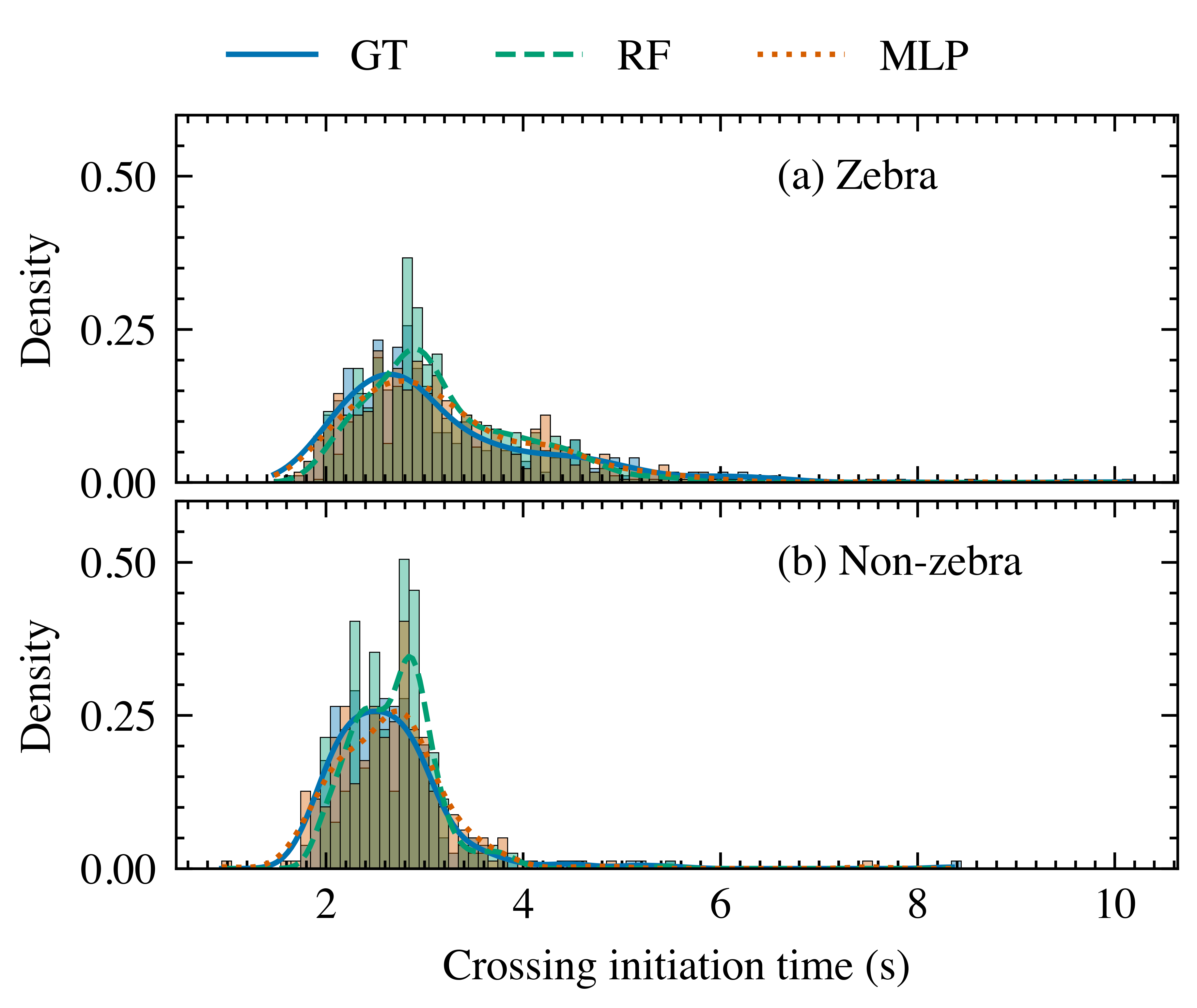}
    \caption{Distribution of predicted crossing initiation time. The y-axis is the normalized density such that the total area of the histogram in each subplot is equal to 1.}
    \label{fig:cit_distribution}
\end{figure}

We further compare the distribution of predicted CIT of RF and MLP models with groundtruth. The distribution density is shown in Fig.~\ref{fig:cit_distribution}. Although the error of RF is the smallest, the distribution of MLP is closer to the groundtruth, which indicates the neural network model may have better generalizability.

\paragraph{Crossing duration}
The prediction error of CD is shown in Table~\ref{tab:pred_error_cd}. Input features are the same as in crossing decision prediction models. Compared with the baseline LR results, RF and MLP gain great improvements. The MLP model achieves the best results, where the MAE and RMSE are 0.282~s and 0.446~s, respectively, showing an improvement of 35.00\% and 30.14\% compared with the LR model.

\begin{table}[h]
    \caption{Prediction Error of Crossing Duration}
    \label{tab:pred_error_cd}
    \centering
    \begin{tabular}{l|cc}
    \hline  
        Model & MAE [s] & RMSE [s] \\  \hline
        LR (baseline)~\cite{Kalantari2022Who} & 0.434 & 0.638 \\
        RF (ours) & 0.297 & 0.458 \\
        MLP (ours)  & \textbf{0.282} & \textbf{0.446} \\
    \hline
    \end{tabular}
\end{table}

\begin{figure}
    \centering
    \includegraphics[width=0.4\textwidth]{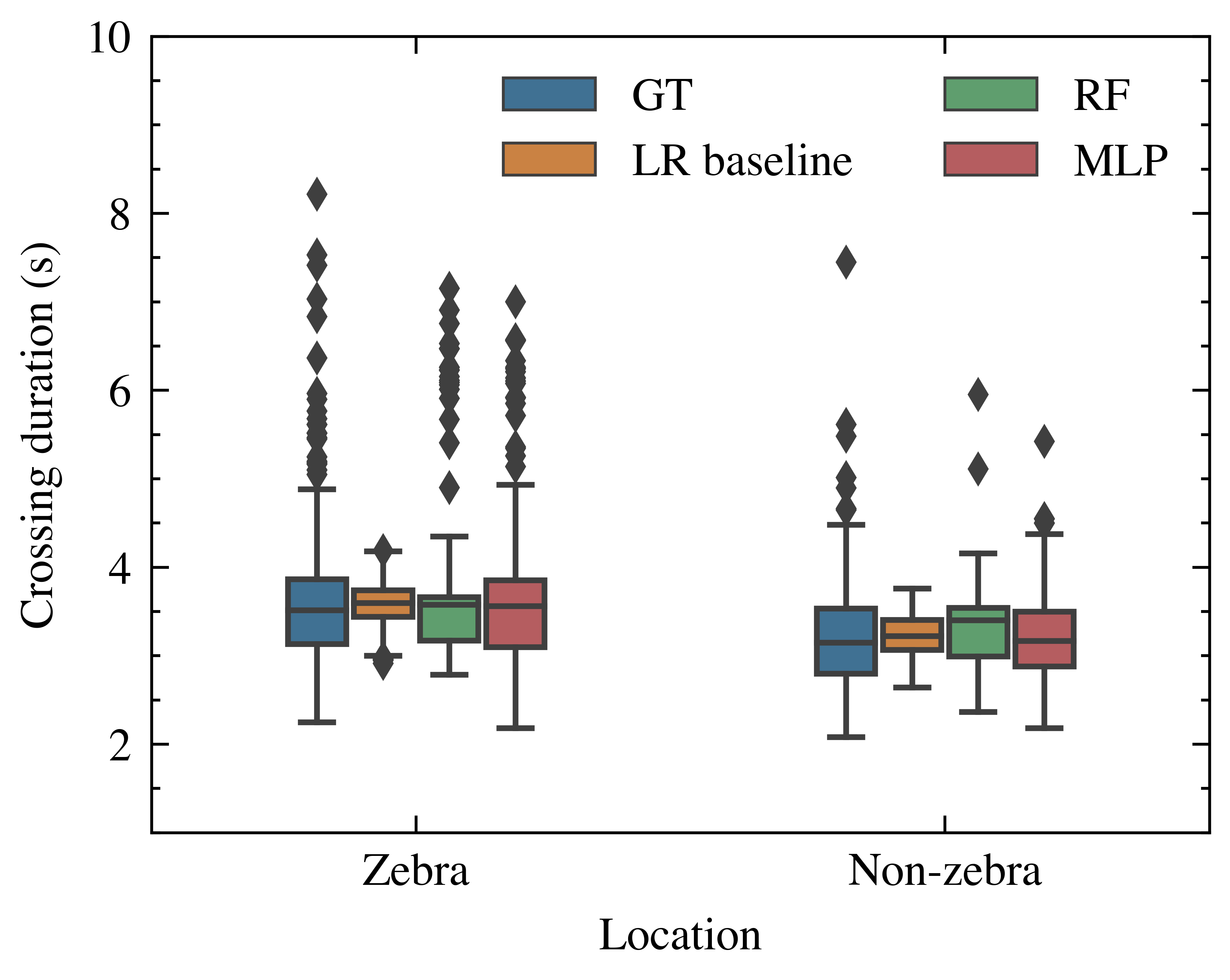}
    \caption{Box plots of crossing duration for GT, linear regression (LR), RF, and MLP models. Outliers are marked as diamond dots.}
    \label{fig:cd_boxplot}
\end{figure}

The box plots of CD for the groundtruth and different models at different crossing locations are shown in Fig.~\ref{fig:cd_boxplot}. Pedestrians who crossed the road at zebra crossings took longer time than those who crossed at non-zebra crossings.
This could be because people feel safer to cross at a zebra crossing so they can take their time. The prediction of the LR baseline model is more concentrated, while the MLP model is more distributed.

\begin{figure}
    \centering
    \includegraphics[width=0.4\textwidth]{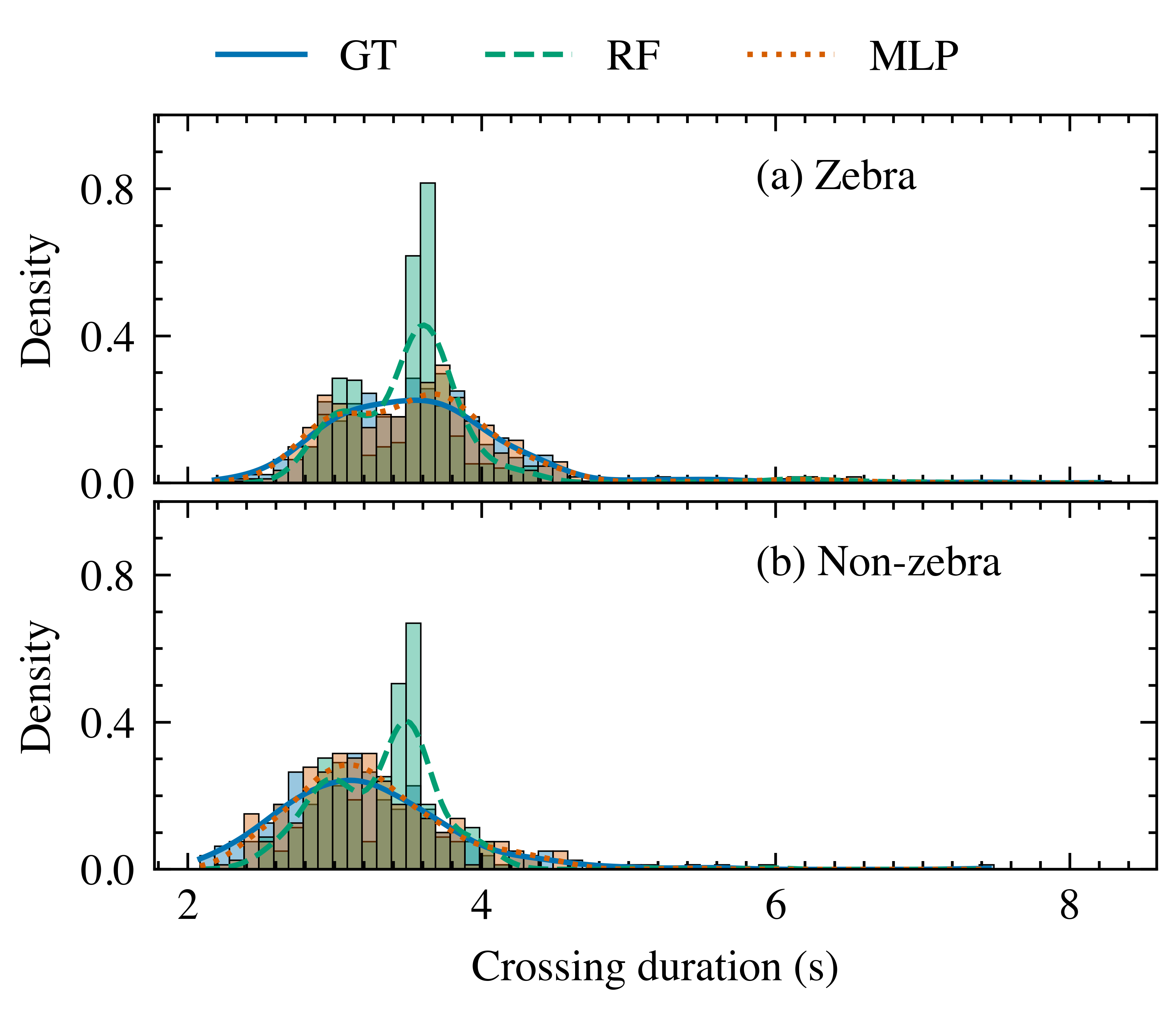}
    \caption{Distribution of predicted crossing duration. The y-axis is the normalized density such that the total area of the histogram in each subplot is equal to 1.}
    \label{fig:cd_distribution}
\end{figure}

The CD distribution density of the groundtruth and RF and MLP predictions are shown in Fig.~\ref{fig:cd_distribution}. The errors of RF and MLP prediction results are close, while the distribution of MLP is shown to be closer to the groundtruth. This indicates that the MLP model is more capable of predicting crossing duration.

\subsection{Ablation Study}
\label{sec:ablation_study}
In practice, it is usually difficult to obtain all the features we use in our models. Therefore, it is necessary to investigate the performance of the predictive models with only parts of the features. We consider four subsets of features as follows.

\textit{Feature subset 1:} $T_{a},T_{w},L,A_d,A_p,G_d,G_p$. Compared with the original feature set, subset 1 removes the features related to the personalities of the drivers and pedestrians, i.e. SVO and AISS information. This feature set fits scenarios without personality traits.

\textit{Feature subset 2:} $T_{a},T_{w},L,A_d,G_d$. Subset 2 removes the age and gender information of pedestrians while keeping this information for drivers. This feature set fits scenarios where the driver's information is known to aid decision-making on a private vehicle for example.

\textit{Feature subset 3:} $T_{a},T_{w},L,A_p,G_p$. Subset 3 removes the age and gender information of drivers while keeping this information for pedestrians. This feature set fits scenarios where the pedestrian's age and gender information is known, for example, for the predictions on a labeled publicly available dataset.

\textit{Feature subset 4:} $T_{a},T_{w},L$. Subset 4 only keeps TTA, waiting time, and crossing location type. This is the basic feature set that the information is easy to get, and the features are important for all predictive models.


\paragraph{Crossing decision} The crossing decision prediction results using different subsets of features are shown in Table~\ref{tab:pred_acc_subset}. Surprisingly, with fewer input features, the LR model does not get a significant drop in performance, while the RF and MLP models get much worse results compared with using all features. This shows that the LR model depends mainly on several important basic features for prediction, while the other two models rely on all features.

For subset 1, the MLP model gets the best results with accuracy and F1 score of 88.43\% and 91.30\%, respectively. The neural networks are able to handle the case when there are many features with non-linear relations. Without SVO and AISS information, the performance of MLP drops, so they are important for the MLP model. For subsets 2 and 3, the RF model gets the best performances, while MLP gets the worst results. This indicates that age and gender features play important roles in the MLP model. As discussed in Sec.~\ref{sec:results_crossing_decision}, the age and gender features are not in the top six important features of the RF model. Therefore, even without age and gender information, it can still provide a relatively good prediction compared with the other two models. For subset 4, the RF model gets the best performance, while the LR model achieves competitive results. This indicates that for a simple model with very few input features, the linear model is still strong and powerful compared with the other two non-linear models.

\begin{table}[h]
    \caption{Crossing Decision Prediction Using a Subset of the Features}
    \label{tab:pred_acc_subset}
    \centering
    \begin{tabular}{m{1cm}|m{0.7cm}m{0.8cm}|m{0.7cm}m{0.8cm}|m{0.8cm}m{0.7cm}}
    \hline
        Model & \multicolumn{2}{c|}{LR} & \multicolumn{2}{c|}{RF} & \multicolumn{2}{c}{MLP} \\ \hline
        Features & ACC & F1 & ACC & F1 & ACC & F1 \\  \hline
        All & 85.93\% & 89.32\%	& 89.84\% & 92.44\% & \textbf{90.23\%} & \textbf{92.47\%} \\
        SubSet 1 & 85.77\% & 89.24\% & 87.89\% & 90.90\% & \textbf{88.43\%} & \textbf{91.30\%} \\
        SubSet 2 & 85.38\% & 88.97\% & \textbf{86.09\%} & \textbf{89.54\%} & 83.35\% & 87.38\% \\
        SubSet 3 & 85.93\% & 89.34\% & \textbf{86.56\%} & \textbf{89.92\%} & 83.19\% & 87.26\% \\
        SubSet 4 & 85.54\% & 89.15\% & \textbf{85.77\%} & \textbf{89.36\%} & 85.69\% & 89.35\% \\
    \hline
    \end{tabular}
\end{table}

\paragraph{Crossing initiation time} The prediction results for CIT using subsets of features are shown in Table~\ref{tab:pred_cit_subset}. With fewer input features, the errors of LR models are increased slightly, and the errors of RF and MLP models are increased to a great extent. This indicates that the LR model relies more on TTA, waiting time, and crossing location type, while the other two non-linear predictive models also depend on personality trait features, age, and gender.

\begin{table}[h]
    \caption{Crossing Initiation Time Prediction Using a Subset of the Features}
    \label{tab:pred_cit_subset}
    \centering
    \begin{tabular}{m{1cm}|m{0.7cm}m{0.8cm}|m{0.7cm}m{0.8cm}|m{0.8cm}m{0.7cm}}
    \hline
        Model & \multicolumn{2}{c|}{LR} & \multicolumn{2}{c|}{RF} & \multicolumn{2}{c}{MLP} \\ \hline
        Features & MAE & RMSE & MAE & RMSE & MAE & RMSE \\  \hline
        All & 0.616 & 0.900 & \textbf{0.428} & \textbf{0.704} & 0.500 & 0.794 \\
        Subset 1 & 0.646 & 0.945 & \textbf{0.485} & \textbf{0.757} & 0.524 & 0.802 \\
        Subset 2 & 0.665 & 0.959 & \textbf{0.542} & \textbf{0.817} & 0.679 & 0.999 \\
        Subset 3 & 0.644 & 0.948 & \textbf{0.558} & \textbf{0.823} & 0.618 & 0.901 \\
        Subset 4 & 0.678 & 0.970 & \textbf{0.659} & \textbf{0.949} & 0.694 & 0.980\\
    \hline
    \end{tabular}
\end{table}


\paragraph{Crossing duration} The prediction results for CD using subsets of features are shown in Table~\ref{tab:pred_cd_subset}. With fewer input features, the errors of LR models slightly increase. The other two non-linear models show a great increase in errors with fewer features. MLP performs best with all features and performs worst with subset 4. This indicates that MLP is more suitable for prediction with a large number of input features while less capable for prediction with fewer features.

\begin{table}[h]
    \caption{Crossing Duration Prediction Using a Subset of the Features}
    \label{tab:pred_cd_subset}
    \centering
    \begin{tabular}{m{1cm}|m{0.7cm}m{0.8cm}|m{0.7cm}m{0.8cm}|m{0.8cm}m{0.7cm}}
    \hline
        Model & \multicolumn{2}{c|}{LR} & \multicolumn{2}{c|}{RF} & \multicolumn{2}{c}{MLP} \\ \hline
        Features & MAE & RMSE & MAE & RMSE & MAE & RMSE \\  \hline
        All & 0.428 & 0.615 & 0.297 & 0.458 & \textbf{0.282} & \textbf{0.446} \\
        SubSet 1 & 0.474 & 0.668 & 0.345 & 0.496 & \textbf{0.321} & \textbf{0.478} \\
        SubSet 2 & 0.478 & 0.677 & \textbf{0.418} & \textbf{0.563} & 0.454 & 0.618 \\
        SubSet 3 & 0.472 & 0.668 & \textbf{0.418} & \textbf{0.613} & 0.507 & 0.708 \\
        SubSet 4 & \textbf{0.476} & \textbf{0.677} & 0.491 & 0.685 & 0.502 & 0.702 \\
    \hline
    \end{tabular}
\end{table}

\section{Conclusion}
In this paper, we have proposed predictive models for pedestrian crossing behavior when interacting with vehicles at unsignalized crossings. Our proposed MLP predictive model has improved the prediction accuracy and F1 score of crossing decisions by 4.46\% and 3.23\% compared to that of the logistic regression baseline. For crossing initiation time prediction, the proposed RF model decreases the RMSE error by 21.56\%, and for crossing duration, the MLP model reduces the RMSE error by 30.14\%. Furthermore, we have analyzed the importance and influence of the various input features in different models. The presence of zebra crossing, TTA, AISS of pedestrians, and pedestrian waiting time are important features for all models for crossing decision prediction. Additionally, we have applied an ablation study and presented the predictability of crossing behavior with fewer input features. This provides guidance for selecting proper models when there are only limited input features.
In future work, more information such as pedestrians' trajectories and postures can be included to support building predictive models.







%
\bibliographystyle{IEEEtran}
\bibliography{IEEEexample}

\end{document}